\makeatletter\def\graphicscache@inhibit{true}\makeatother
\documentclass[keeplastbox, letterpaper, 10 pt, conference]{ieeeconf}

\IEEEoverridecommandlockouts                              %

\usepackage{times}

\usepackage{graphics} %
\usepackage{graphicx}
\usepackage{todonotes}
\usepackage{units}

\usepackage[
	style        = ieee,
	natbib       = true,
	backend      = bibtex,
	firstinits,
	doi          = false,
	mincitenames = 1,
	maxcitenames = 2,
	maxbibnames  = 99,
	sorting      = none,
	terseinits   = false,
	hyperref     = true]{biblatex}
\usepackage{multicol}
\usepackage[bookmarks=true]{hyperref}

\usepackage[bottom]{footmisc}		%

\usepackage{amssymb}     %
\usepackage{balance}
\usepackage{amsmath}     %
\usepackage{booktabs}
\usepackage{multirow}

\usepackage{graphicscache}

\newcommand\norm[1]{\left\lVert#1\right\rVert}

\hypersetup{
	pdfinfo={
	   Author={Diego Rodriguez},
	   Title={DeepWalk: Omnidirectional Bipedal Gait by Deep Reinforcement Learning},
	   Subject={Robotics},
	   Keywords={Robotics;DeepReinforcementLearning,Walking,Gait}
	}
}

\begin{document}

	\title{\LARGE \bf DeepWalk: Omnidirectional Bipedal Gait by\\ Deep Reinforcement Learning}

\author{Diego Rodriguez and Sven Behnke%
	\thanks{Both authors are with the Autonomous Intelligent Systems (AIS) Group, Computer Science Institute VI, University of Bonn, Germany {\tt\small rodriguez@ais.uni-bonn.de}}
}%

\maketitle
\thispagestyle{empty}
\pagestyle{empty}

\begin{tikzpicture}[remember picture,overlay]
\node[anchor=north west,align=left,font=\sffamily,yshift=-0.2cm,xshift=3.5cm] at (current page.north west) {%
	In: Proceedings of the International Conference on Robotics and Automation (ICRA) 2021
};
\end{tikzpicture}%

\begin{abstract}
Bipedal walking is one of the most difficult but exciting challenges in robotics.
The difficulties arise from the complexity of high-dimensional dynamics,
sensing and actuation limitations combined with real-time and computational constraints.
Deep Reinforcement Learning (DRL) holds the promise to address these issues by fully exploiting the robot dynamics with minimal craftsmanship.
In this paper, 
we propose a novel DRL approach that enables an agent to learn omnidirectional locomotion for humanoid (bipedal) robots.
Notably, 
the locomotion behaviors are accomplished by a single control policy (a single neural network).
We achieve this by introducing a new curriculum learning method that gradually increases the task difficulty by scheduling target velocities.
In addition, 
our method does not require reference motions which facilities its application to robots with different kinematics,
and reduces the overall complexity.
Finally, 
different strategies for sim-to-real transfer are presented which allow us to transfer the learned policy to a real humanoid robot.
\end{abstract}

\IEEEpeerreviewmaketitle

\section{Introduction}
\label{sec:intro}
Humanoid robots are one of the most versatile and flexible platforms for acting in made-for-human environments.
This versatility comes, however, at the cost of complexity.
Bipedal locomotion poses still several challenges in terms of planning and control mainly due to the high dimensionality,
complex dynamics, sensing and actuation limitations, and real-time constraints. 
Inspired by the human example,
Deep Reinforcement Learning (DRL) approaches offer a promising model-free alternative to address these issues by making use of prior experiences.

Several state-of-the-art DRL-based locomotion controllers employ tracking-based policies of reference motions to learn separate controllers to walk at fixed velocities~\cite{peng2017deeploco,xie2018feedback,xielearning,tan2018}.
In contrast,
we propose a single policy in order to avoid training and combining separate control policies.
The parameter sharing facilitates information transfer when learning different velocities.
We further reduce complexity by removing the need for motion capture data and engineered modules such as kinematic mappings and motion interpolations.
This is made possible by the introduction of a nominal pose that can be defined as the standing pose of the robot.

\begin{figure}[t]
	\centering
	\includegraphics[width=1.0\linewidth]{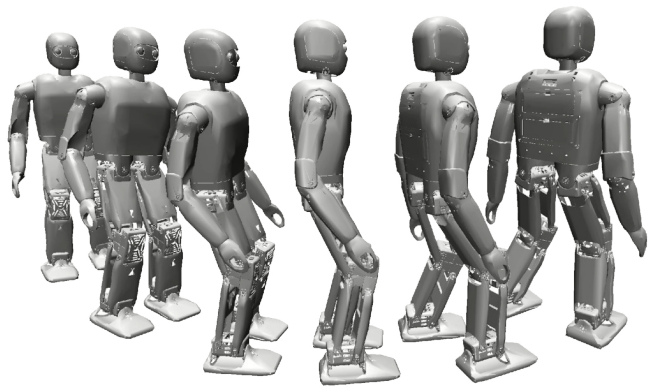} 
	\caption{Omnidirectional walk executed by our learned control policy. 
		The robot is able to turn left during forward walking.}
	\label{fig:teaser}
	\vspace*{-3ex}
\end{figure}

The core contribution of this paper is a novel approach to learn a \textit{single} control policy for \textit{omnidirectional} walking using DRL on a realistic humanoid robot model without using reference motions.
To achieve this, 
we propose a velocity scheduler that gradually increases the task difficulty of the agent, 
and we introduce a nominal pose to guide the learning process.
We additionally introduce a new way of controlling the use of motor power (torque) by bounding the action space through beta policies~\cite{chou2017improving}.

Our experiments demonstrate that the learned policy can successfully produce omnidirectional motions allowing the robot to walk forwards, backwards, laterally, diagonally, to turn around the vertical axis and to combine these directions.
Moreover, 
our learned controller is successfully transferred to real hardware.
\section{Related Work}
\label{sec:related}

Learning approaches applied to walking controllers have been predominantly investigated for gait optimization~\cite{Rodriguez2018Combining,rai2018,farazi2013evolutionary}.
These methods,
however,
depend on engineered components such as trajectory planners, 
central pattern generators and robot dynamics models.
Recently, 
model-free locomotion controllers have been developed by the character animation community based on DRL algorithms~\cite{heess2017emergence,xie2020,peng2017deeploco,peng2018deepmimic,bergamin2019drecon,park2019learning}.
\citet{heess2017emergence} generated robust locomotion maneuvers for different characters by applying curriculum strategies on the environment.
In \cite{xie2020},
3D walking motions were achieved by employing curriculum strategies on the commanded velocities.
Still, 
the policies were trained on animation characters and their applicability in real robots is questionable.
\citet{peng2017deeploco} 
provide human motion capture data with desired foot-placement goals to guide the reinforcement learning process.
\citet{peng2018deepmimic} later proposed DeepMimic,
an approach that is able to mimic highly dynamic motions such as backflips, cartwheels, and rolls.
Inspired by DeepMimic, 
\citet{bergamin2019drecon} and \citet{park2019learning} learned kinematic trajectories which are tracked by dynamic-consistent RL-based tracking controllers.
However, in all these approaches, 
the controllers were demonstrated on simplified models with high torque capabilities and ground truth data from the simulator which do not resemble real-world conditions.

Deep reinforcement learning has also been applied in the robotics community~\cite{hwangbo2019learning, yang2018learning, xie2018feedback, xielearning, tsounis2019deepgait, yu2019sim}.
\citet{hwangbo2019learning} presented a control policy which is later transferred to a real quadruped robot.
The transfer was possible thanks to a network trained to model the dynamics of the actuators. 
In contrast to~\cite{hwangbo2019learning}, 
our approach employs a much simpler reward function and it is demonstrated in a humanoid robot which imposes a harder balance problem.

One of the first attempts to apply DRL on humanoid robots was proposed by~\citet{yang2018learning},
who integrated the capture-step equations into the reward function for learning push-recovery capabilities.
However,
walk capabilities were not developed.
\citet{xie2018feedback} learned a forward walk with a biped robot using DRL based on separate reference motions at different velocities.
Interpolation between policies needs to be explicitly handled to allow the robot to change between commanded velocities.
Later,
\citet{xielearning} extended this approach and transferred the policies to a real platform.
Similarly, 
\cite{yu2019sim} transferred a walking controller into a real bipedal robot.
In both approaches, 
the achieved locomotion capabilities rely on multiple control policies that need to be trained separately.
Our approach, 
on the other hand, 
is able to generate an omnidirectional gait using a single policy without any reference motion.

\section{Background}
\label{sec:background}

\subsection{Deep Reinforcement Learning}
\label{sec:deept_rl}

Reinforcement learning algorithms aim to find a policy $\pi$ based on experiences that will guide an agent to solve a specific task.
This task is modeled as a Markov Decision Process (MDP) defined by the tuple $\{S, \mathcal{A}, P, \gamma, r\}$,
where $S\in\mathbb{R}^n$ represents the state space,
$\mathcal{A}\in\mathbb{R}^m$ is the set of actions the agent can take,
$P:S\times \mathcal{A}\mapsto S$ models the dynamics of the system,
$\gamma\in[0,1]$ is a discount factor,
and $r:S\times \mathcal{A}\mapsto \mathbb{R}$ is a reward function that rewards or punishes an action $a_t$ taken in state $s_t$ after interacting with the environment at time step $t$.

In this paper, 
we focus on model-free reinforcement learning,
and we directly construct a parametrized policy $\pi_\theta$ by maximizing a cost function $J(\theta)$ with respect to the parameters $\theta$, 
without explicitly modeling the dynamics $P$.
As common in continuous control problems, 
we define $\pi_\theta$ as a stochastic policy $\pi_\theta(a|s)$ defined as a probability distribution of taking an action $a_t$ given a state $s_t$.
Policy gradient algorithms try to solve this optimization problem by sampling trajectories around the current policy $\pi_\theta$ and by updating the parameters $\theta$ according to the gradient $\nabla_\theta J(\theta)$ in an ascent fashion.
This gradient is expressed as:
\begin{equation}
\nabla_\theta J(\theta) = \nabla_\theta\mathbb{E}\left[\Psi\right] = \mathbb{E}\left[ \Psi_t\nabla_\theta\log\pi_\theta(a_t|s_t)\right].
\label{eq:gradient}
\end{equation}
Eq.~(\ref{eq:gradient}) tells us how the parameters $\theta$ should be updated judged by the score function $\Psi_t$
which can take several forms including: 
the state-action value function, 
		$Q^\pi(s_t,a_t)$, %
	the advantage function,
		$A^\pi(s_t,a_t)$, %
and the Generalized Advantage Estimator (GAE): 
\begin{equation}
	A^{GAE(\gamma,\lambda)}=\sum_{l=0}^{\infty}(\gamma\lambda)^l\delta_{t+l}^{V},
	\label{eq:gae}
\end{equation}
that reduces the variance of the gradient estimates at the cost of introducing bias~\cite{schulman2015high}.
This is done by combining a series of Temporal Difference (TD) residuals $\delta_t^V=r_t+\gamma V(s_{t+1})-V(s_t)$ through a parameter $\lambda\in[0,1]$ that trades off variance and bias.
The value function is defined as $V^\pi(s_t) = \mathbb{E}_{s_{t+1},a_{t}}\left[\sum_{l=0}^\infty r_{t+l}\right]$.

Estimators of the gradient policy can additionally be obtained by automatic differentiation of an objective function constructed such that its gradient is the policy gradient estimator, e.g.:
\begin{equation}
	L(\theta) = \mathbb{E}_t\left[ A_t\log \pi_\theta(a_t|s_t) \right].
\end{equation}
Alternatively, a surrogate objective function can be used, 
as introduced in the Proximal Policy Optimization (PPO) algorithm~\cite{schulman2017proximal}. 
This surrogate has the form:
\begin{equation}
	L^{PPO}(\theta) = \mathbb{E}_t\left[ \min\left(r_t(\theta)A_t,\textrm{clip}\left( r_t(\theta),1-\epsilon,1+\epsilon\right)A_t\right) \right],
	\label{eq:L_ppo}
\end{equation}
where, $r_t=\frac{\pi_\theta(a_t|s_t)}{\pi_{\theta_{old}}(a_t|s_t)}$ denotes the probability ratio, 
and $\epsilon$ is a hyperparameter that defines a range in which the new policy is allowed to differ from the previous one.
\section{Method}
\label{sec:method}
\begin{figure}
	\centering
	\includegraphics[width=1.0\linewidth]{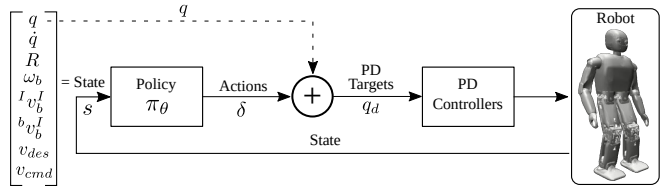} 
	\caption{Control system overview. According to the state $s$ of the robot,
		 the control policy $\pi_\theta$ calculates increments $\delta$ of the current joint state positions $q$ that define targets $q_d$ for PD controllers of the robot. }
	\label{fig:overview}
	\vspace*{-2ex}
\end{figure}

We propose a novel approach to learn an omnidirectional walking controller for humanoid robots.
This controller is parametrized by a neural network called the policy (actor) network (Sec.~\ref{sec:network}).
An overview of the walking controller is presented in Fig.~\ref{fig:overview}.
Our approach consists of two phases: training and inference.
In the training phase, two networks (actor and critic) are learned through experiences that the agent collects acting in the environment using the current parameters of the networks.
Once enough experiences have been collected, the networks' weights are updated and the experience rollout buffers are cleared out.
In the inference phase, the network weights are kept fixed.
For brevity,
the dependency on $t$ will be dropped.
Given the observed state $s$ of the robot, 
the network outputs offsets $\delta$ that are added to the current joint positions $q$ that ultimately define targets $q_d$ for PD controllers of the robot joints.

\subsection{State Space}
\label{sec:state_space}

We define the state $s$ of the robot as: joint positions $q$,
joint velocities $\dot{q}$,
orientation of the base of the robot $R$,
angular velocity of the base $\omega_b$,
velocity of the base w.r.t. a inertial reference frame $I$ expressed in $I$ ${}^{I}v_b^{I}$~\footnote{~Notation: 
	for a vector ${}^Av_C^B$, the left superscript denotes that the coordinates of the vector are expressed in frame $A$, 
	for the position of a point $C$ relative to other point $B$. 
} and in the base link frame ${}^{b}v_b^{I}$,
long-term desired velocity $v_{des}$,
and short-term commanded velocity $v_{cmd}$.

The joint positions $q\in\mathbb{R}^{n_q}$ and velocities $\dot{q}\in\mathbb{R}^{n_q}$ are read directly from the joint encoders,
for a robot with $n_q$ fully actuated joints.
$R$ is a vector containing the orientation of the base link (roll $R_\alpha$, pitch, $R_\beta$ and yaw $R_\gamma$).
Robots without yaw estimation will define a two-dimensional $R$ vector.

The angular velocity ${\omega_b=[\omega_x,\omega_y,\omega_z]\in\mathbb{R}^{3}}$ is taken from the gyro measurements.
The linear velocity of the base link $v_b$ is estimated using the robot kinematics and orientation assuming a flat ground.
The reference frame $I$ initially equals the base link frame and is updated periodically to a frame placed in the base link aligned with the $z$ world axis.
The linear velocity is expressed in the reference frame ${}^{I}v_b^{I}\in\mathbb{R}^{2}$
and in the local frame ${}^{b}v_b^{I}\in\mathbb{R}^{2}$,
and both are included into the state $s$.
Without ${}^{I}v_b^{I}$ in $s$, 
the agent would not be penalized for deviating from the original direction of the desired velocity;
and without ${}^{b}v_b^{I}$,
the agent is susceptible to develop locomotion patterns that follow the walking direction,
but infringe the desired relative velocity.

Finally, the state also incorporates the desired velocity $v_{des}=[v_x^{des},v_y^{des},\omega_z^{des}]\in\mathbb{R}^{3}$ and an immediately commanded velocity $v_{cmd}=[v_x^{cmd},v_y^{cmd},\omega_z^{cmd}]\in\mathbb{R}^{3}$.
The former can be seen as the task's goal.
It represents the user input for controlling the robot.
The latter is an interpolated velocity between the previous desired velocity $v_{des}^{old}$ and the new desired velocity $v_{des}$ to have smoother transitions when changing velocities.

\subsection{Action Space}
\label{sec:action_space}
The actions are represented as a delta $\delta$ that needs to be added to the current joint position $q$, 
such that PD targets are formulated as $q_d=q+\delta$.
This formulation is based on the observation that abrupt changes in the PD target saturate the PD controllers producing jerky,
unnatural high-torque motions.
This issue is addressed by putting limits on the values of $\delta$,
i.e., we respect actuator speed limits.
In stochastic continuous control, 
the commonly used Gaussian policies are not able to handle bounded action spaces like the one proposed here.
\citet{chou2017improving} introduced Beta policies, 
which can deal with bounded action spaces and have shown faster convergence and higher scores than the Gaussian ones.
In our method, we follow Beta policies and bound the action space such that $\delta\in[\delta_{min},\delta_{max}]$.
The actions are sampled from the beta distribution at \unit[40]{Hz}.

\subsection{Reward Function}
\label{sec:reward}
The reward function is defined as: 
\begin{align}
	r=w_v r_{vel}+w_r r_{reg}+w_a r_{alive}+w_f r_{foot}.
\end{align}

All the reward terms are bounded such that $r_*\in[0,1]\cdot\Delta t$, 
where $\Delta t$ is the time step of the policy controller. 
To bound the reward terms,
we use the smooth logistic kernel function $K:\mathbb{R}\rightarrow[0,1]$ expressed as $K(x|l)=2/(e^{lx}+e^{-lx})$,
where $l$ defines the sensitivity of the kernel.
The reward consists of the following terms:

\subsubsection{Velocity Tracking}
\label{sec:rew_vel}
This term states how good the velocity is tracked by the control policy.
$r_{vel}$ is defined as:
\begin{align}
r_{vel}=C_vK(e_v)\Delta t, \;
e_{v} = \norm {
	\begin{bmatrix}
	v_x^{cmd} - {}^{b}v_{b,x}^{I}\\
	v_y^{cmd} - {}^{b}v_{b,y}^{I}\\
	v_x^{cmd} - {}^{I}v_{b,x}^{I}\\
	v_y^{cmd} - {}^{I}v_{b,y}^{I}\\
	\omega_z^{cmd} - \omega_z
	\end{bmatrix}
}.
\end{align}
The tracking error $e_v\in\mathbb{R}^5$ includes the difference between the commanded velocity $v^{cmd}$ and the %
velocity of the base of the robot,
expressed in the reference ${}^{I}v_{b}^{I}$ and the local frame ${}^{b}v_{b}^{I}$,
and the angular velocity error  $\omega_z^{cmd} - \omega_z$.
Note that the L2-norm encourages improvement of all components together.
The value of $C_v\in(0,1]$ changes dynamically the priority of this reward term.
At the beginning of training,
the priority is low to let the agent learn to stand.
The priority is increased rapidly once the robot has a notion of standing.
The value of $C_v$ is defined per epoch: $C_{t+1}=C_t^{k_d}$,
where $k_d$ specifies the speed change of $C_v$.
Without $C_v$, 
the agent would learn a greedy policy in which it tilts to the front,
provoking a fall.

\subsubsection{Pose Regularization}
\label{sec:rew_reg}
The regularization of the learning process in our approach is done through a single joint position configuration.
This nominal pose can be defined, 
for example, 
as the standing pose of the robot.
The regularization error is defined as the difference between the nominal pose $q^{req}$ and the corresponding joint positions to regularize $q^r$:
\begin{align}
	r_{reg}=K(\norm{q^{reg}-q^r})\Delta t.
\end{align}
Note that not all joints in $q$ need to be regularized.

\subsubsection{Alive}
\label{sec:rew_alive}
The agent is referred to be alive if it is not in the process of falling.
In each step, a fixed reward is given if the height of the base with respect to the floor, $p_z$, remains above a defined value, $p_z^{min}$, or if the roll $R_\alpha$ and pitch $R_\beta$ angles of the base stay below thresholds, $R_\alpha^{max}$, and $R_\beta^{max}$, respectively.
In case these thresholds are violated,
a fall is expected and the rollout is terminated.
In contrast to a large negative reward given at the terminal state,
$r_{alive}$ is proportional with the length of the rollout such that longer sequences are more rewarded,
i.e., the agent is encouraged to learn keeping balance.
$r_{alive}$ is formulated as:
\begin{align}
r_{alive} = 
\begin{cases}
\Delta t &\quad\text{if } p_z>p_z^{min}, R_{\alpha}<R_\alpha^{max}, R_{\beta}<R_\beta^{max}\\
0 &\quad\text{else}.
\end{cases}
\end{align}

\subsubsection{Foot Clearance}
Without a foot clearance term,
policies might be learned which lift the feet as little as possible.
Although such policies show stable walking patterns in simulation,
transferred policies to the real robot exhibit,
in general,
motions with dragging feet partly caused by model differences and joint backlash.
The foot clearance term acts on the swing leg only.
Apart from considering the clearance $^{lf,rf}p_{z}$,
this term includes the roll angle $^{rf,lf}\phi_{x}$ of the right ($rf$) and left ($lf$) foot to discourage the agent to walk on its lateral feet edges which emerges as an artifact from maximizing $^{lf,rf}p_{z}$.
For a right swing leg,
for instance, 
the foot clearance reward term is formulated as:
\begin{gather}
 	 r_{foot} = C_{f}K(e_{f})\Delta t, \nonumber\\
	 e_{f}=\norm{
		\begin{bmatrix}
			w_{\phi}({}^{rf}p^{des}_{z} - {}^{rf}p_{z}),
			{}^{rf}\phi_{x}, 
			{}^{lf}\phi_{x}, 
			w_{\phi}{}^{lf}p_{z}
		\end{bmatrix}^T
	}.
	\label{eq:deep_walk:clearance}
\end{gather}
The minimization of the clearance of the foot in stance disfavors the development of flight phases,
which produce unstable behaviors when the policy is executed on the real robot.
For a left swing leg,
the superscripts of Eq.~(\ref{eq:deep_walk:clearance}) are interchanged correspondingly.
The swing leg is defined as the leg whose foot has the smallest distance to the trunk link in the $z$ axis.
Hysteresis is added when changing the swing leg to discourage flight phases.

\subsection{Actor and Critic Networks}
\label{sec:network}
In our approach, two networks are trained:
a critic and an actor network.
The critic network is employed to estimate a state value function $V_\phi$ used for calculating the generalized advantage estimator $A^{GAE}$ (Eq.~(\ref{eq:gae})).
The parameters of the critic network are updated by minimizing the loss function:
\begin{align}
L^V=(V_\phi(s_t)-\hat{V})^2,
\end{align}
where $\hat{V}$ refers to the sampled state value from the trajectories.
The parameters of the actor network are updated by maximizing the PPO loss function (Eq.~\ref{eq:L_ppo}).

The critic and actor network architectures are very similar.
They contain two fully-connected hidden layers of 512 units with $\tanh$ activation functions and a fully-connected input layer for the state vector $s_t$.
The last layer of the critic network contains a unit that represents the estimated state value $V_\phi$.
The actor network has in its last layer two units for each dimension of the action space.
These units parametrize the beta distributions of the actions. 
In training, 
the actions are sampled from these distributions whereas in inference the actions are described by the distribution mode.
We train our control policy using PPO combined with GAE.

\subsection{Curriculum Learning for Target Velocities}
\label{sec:curriculum}
We formulate the problem of learning a control policy for omnidirectional walking as a curriculum learning problem~\cite{bengio2009curriculum}.
This curriculum strategy is implemented as a velocity scheduler that increases the task difficulty gradually.
The velocity scheduler defines the bounds from where a target velocity $v^{des}\in\mathbb{R}^3$ is sampled each episode from a three-dimensional uniform distribution.
In the first episode,
the robot is commanded to walk only in the sagittal direction at a fixed velocity $v_{core}$.
The bounds of the regions,
from where the target velocities are sampled, 
are gradually increased as training progresses.
We define an episode number $\zeta$,
in which the target velocities stop increasing,
The bound values for each episode are linearly interpolated according to $\zeta$ and the maximum target velocities. 
The training process continues until the maximum number of episodes is reached.
In this manner, 
the agent can refine the learned policy.

The core velocity $v_{core}\gg[0,0,0]$ is introduced for several reasons: 
first, 
to encourage more walking than standing behaviors,
second, 
to avoid forcing the agent to learn at the beginning to walk forward and backward simultaneously,
which is a harder task compared to walking forward,
and finally, 
to prevent the agent learning to slide instead of walking.

\subsection{Sim-to-real Transfer}
\label{sec:transfer}
In this section,
we introduce several strategies that facilitate the sim-to-real transfer.%

	\subsubsection{System Identification}
	This contributes to finding a good initial set of simulation parameters.
	We found that the reflected inertia is a decisive parameter for learning a stable gait.
	Specifically, 
	we notice that low values of this parameter lead to jerky motions that are very challenging to control.
	
	Tuning the PD controllers of the real robot and the simulation to get similar responses is critical.
	Because the responses of the PD controllers influence the joint state configuration 
	and therefore the network input,
	a single untuned PD controller suffices to produce a previously non-observed input that might lead to continually increasing instabilities.

	\subsubsection{Noise Injection}
	Additional noise is purposely injected to the output network. %
	At the beginning of each episode,
	noise is independently sampled for each joint.
	The noise is sampled from a uniform distribution $\eta_{pd}\sim\mathcal{U}(1-\epsilon_{pd},1+\epsilon_{pd})$.
	$\eta_{pd}$ acts as a scale factor of the inferred target deltas $\delta$.
	The noised PD targets are then defined as $q_d = q + \eta_{pd}\cdotp\delta$.
	To model real sensory data, 
	noise is also added to the sensors (gyro $\eta_g$, accelerometer $\eta_a$, position encoders $\eta_q$ and velocity sensors $\eta_v$).
	The added sensory noises are independently randomly sampled from zero-mean Gaussian distributions. %
	
	\subsubsection{Dynamics Randomization}
	Parameters such as masses, inertia and joint positions are obtained directly from the CAD model.
	For the sim-to-real transfer, 
	the friction model plays a key role,
	and it needs to consider tangential and torsional forces.
	The friction between two surfaces is modeled by elliptic cones.
	The friction coefficients (tangential $\mu_{t}$ and torsional $\mu_{z}$) are randomly sampled from uniform distributions $\mu_{t,z}\sim\mathcal{U}(\mu^{min}_{t,z},\mu^{max}_{t,z})$ at the beginning of each episode.

\begin{figure*}[]
	\centering
	\includegraphics[width=\linewidth]{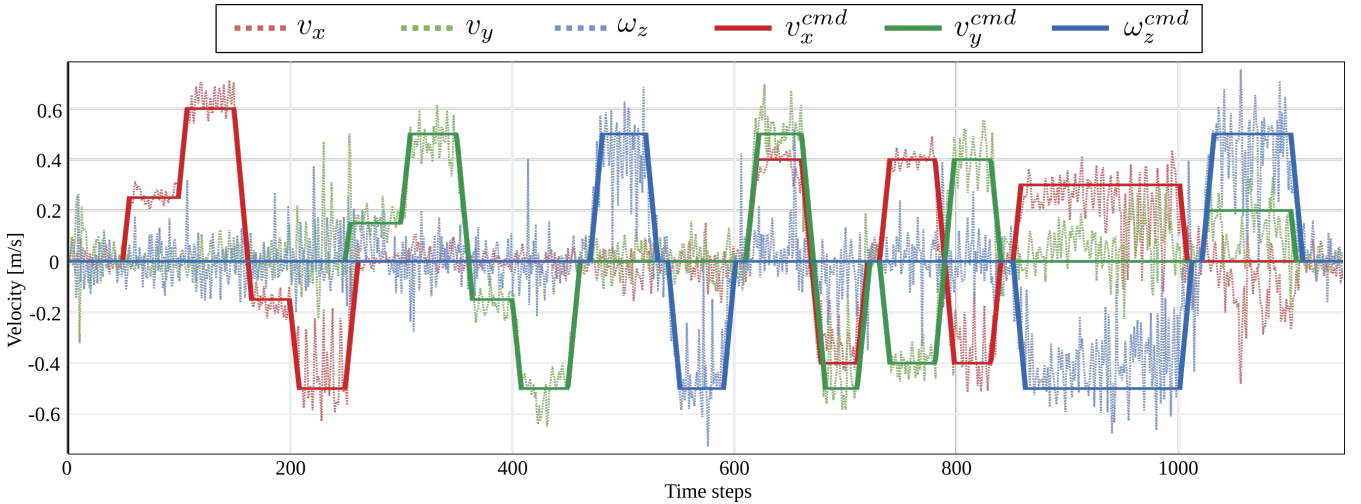} 
	\caption{Velocity tracking. The learned control policy is able to follow the commanded velocities (solid wider lines) in different directions with varying velocities.
		Initially, motions in single directions (sagittal, lateral and turning) are demonstrated.
		After 600 time steps, motions with combined directions (sagittal with lateral, sagittal with turning and lateral with turning) are evaluated. }
	\vspace*{-2ex}
	\label{fig:vel_tracking}
\end{figure*}
	
	\subsubsection{Modeling of Actuation Latency}
	We define latency as the time an actuator takes to read its actual position after a commanded target has been sent.
	In simulation,
	this happens immediately.
	However, 
	with the real robot,
	this delay implies that the learned policy is taking actions on states that do not fully represent the actual state of the robot,
	which leads to unstable gaits and jerky motions.
	Following the approach proposed by~\citet{tan2018},
	we model actuation latency by keeping a history of observations and by feeding previous observations to the network according to a delay time $t_{lat}$.
	Observations are recorded at the frequency of the control policy.
	The network input is then defined as a linear interpolation between adjacent observations according to the latency $t_i\le t_{lat}\le t_{i+1}$.
	To make the policy robust against actuation latency,
	a latency value $t_{lat}$ is uniformly sampled $t_{lat}\sim\mathcal{U}(t^{min}_{lat},t^{max}_{lat})$ at the beginning of each training episode.

	\subsubsection{Network Output Filtering}
	We filter the actions inferred by the policy before sending the corresponding commands to the actuators using a Butterworth low-pass filter.
\section{Evaluation}
\label{sec:results}

Our approach is evaluated on the NimbRo-OP2X robot~\cite{Ficht2018}.
The robot is \unit[135]{cm} high and weights \unit[19]{kg}.
The platform possesses 18 Degrees of Freedom (DoF): 
five on each leg, three on each arm and two for the head.
The joints at the head (pitch and yaw) are neglected because their contribution for walking is considered insignificant and they are needed for active visual perception.
Each leg exhibits a parallel kinematic structure containing: a hip yaw, a hip roll, a hip pitch, a knee pitch and an ankle roll.
For state estimation, the robot uses only a gyroscope and an accelerometer enabling a 2\,DoF state estimation,
namely the roll and the pitch angles.
The action space was bounded to $\delta\in[-0.1, 0.1]$.
In total, our robot creates a 47-dimensional state space and a 16-dimensional action space.

The training is carried out in the physics-based simulator MuJoCo. %
The simulation runs at \unit[1]{kHz}.
The PD controllers have the same frequency as the simulator.
The task was implemented as an OpenAI Gym environment~\cite{openai_gym}.
To speed up the training,
12 parallel environments were running simultaneously on an Intel i9-9900K CPU.
The weights of the reward function terms (${w_v=42, w_r=4}$, $w_a=4$ and ${w_f=18}$) were set manually from experience.
The kernel sensitivities were set to $l_v=9$, $l_r=3$ and $l_f=10$.

The networks are trained iteratively by epochs. %
Each epoch comprises 800 time steps.
Both networks use a learning rate equal to $1\times 10^{-4}$ with a batch size of 480.
Per epoch, 10 updates are performed using the Adam optimizer.
Finally, the decaying factor $\gamma$ is set to 0.99 and $\tau = 0.97$.
The training finished after 7400 epochs for a total of $7.1\times10^7$ time steps,
resulting in 20.5 days of simulated time which corresponds to 32.5 hours of computation in real time.

The robot is asked to learn to walk at  $v_{core}=[\unit[0.4]{m/s}, \unit[0.0]{m/s}, \unit[0.0]{rad/s}]$.
The minimum and maximum velocities in each direction are bounded to $v_x\in[-0.6, 0.6]$, $v_y\in[-0.6, 0.6],$ and $\omega_z\in[-0.6, 0.6]$.
Finally,
the curriculum variables of the reward terms are initialized at $C_v=0.01$ with $k_d=0.95$, and $C_{foot}=0.05$ with $k_d=0.995$.

\begin{figure}[b!]
	\vspace*{-2ex}
	\centering
	\includegraphics[width=\linewidth]{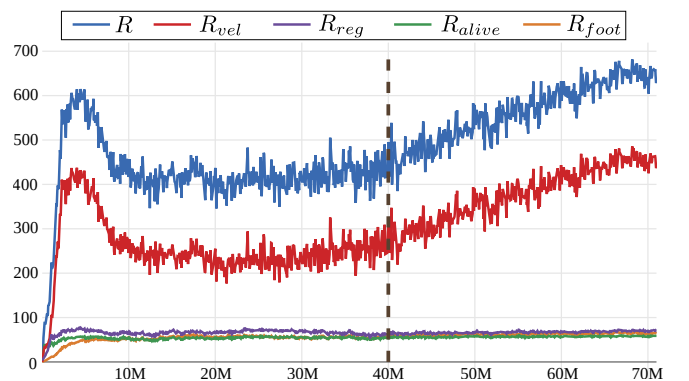} 
	\caption{Training return curves.
		The vertical dashed line represents the point where the limits of the velocity scheduler have been reached.
		Note how the agent continues refining the policy after these limits have been reached.
	}
	\label{fig:return}
\end{figure}

The factor noise to the PD controllers $\epsilon_{pd}$ is set to $0.1$.
The standard deviations of the noise applied to the sensory data are:
${\eta_{g}=1\times 10^{-4}}$,
${\eta_{a}=1\times 10^{-4}}$,
${\eta_{q}=1\times 10^{-3}}$,
and $\eta_{v}=1\times 10^{-3}$.
The friction values are sampled from $\mu_{t}\sim\mathcal{U}(0.4,0.8)$ and
$\mu_{z}\sim\mathcal{U}(0.1,0.3)$,
for the tangential,
and torsional friction coefficients.
The injected noise and sampled friction values are considered only in simulation.
The latency $t_{lat}$ is sampled uniformly from $[0,50]$\unit{ms} in training ($t_{lat}=8$\unit{ms} for the real robot).
Finally, 
the cutoff frequency of the low-pass filter is set to \unit[10]{Hz} for the real robot.

Initially,
we evaluate the learned gait in simulation.
Figure~\ref{fig:vel_tracking} shows the velocity of the base given commanded velocities.
Figure~\ref{fig:return} shows the return training curves.
At the end of training,
the robot falls rarely having an alive reward value oscillating around its maximum (80).
Falls happen mostly when robot is commanded to the velocity limits.
Moreover, 
the robot learned to walk at different speeds in different directions.
Interestingly, 
the robot also learns to walk in place, 
i.e., to lift the feet rhythmically without moving in any direction.
The accompanying video shows the acquired locomotion skills.

In order to evaluate the robustness of the learned controller,
we performed ten walking sequences of \unit[60]{s} for different commanded velocities and counted the number of falls.
The results are presented in Table~\ref{table:falling_test}.
The robot was more susceptible to falling when going backwards and turning along the vertical axis.
The agent had less experiences walking backwards compared to the forward counterpart.
Regarding turning,
the robot was requested to move at the target velocity limits which suggests an increment in these limits.
Additionally, 
the capacity of the controller to switch between velocities was evaluated by sampling uniformly 3-dimensional commanded velocities from the limits defined above.
From 160 changes, 
the robot successfully performed 150 establishing a 93.75\% success rate.

\begin{table}
	\centering
	\caption{Results of falling test}
	\vspace*{-2ex}
	\label{table:falling_test}
	\begin{tabular}{cc|cc}
		\toprule
		Commanded Velocity & \# Falls & Commanded Velocity & \# Falls \\
		\midrule
		$v=[0.0,0.0,0.0]$ & 0/10  &	$v=[0.0,0.0,0.6]$ & 1/10 \\
		$v=[0.6,0.0,0.0]$ & 0/10  & $v=[0.0,0.0,-0.6]$ & 3/10 \\
		$v=[-0,6,0.0,0.0]$ & 2/10 & $v=[0.4,0.4,0.0]$ & 0/10 \\
		$v=[0.0,0.6,0.0]$ & 0/10  & $v=[0.4,0.3,0.0]$ & 0/10 \\
		$v=[0.0,-0.6,0.0]$ & 0/10 & $v=[0.0,0.4,0.3]$ & 0/10 \\
		\bottomrule
	\end{tabular}
	\vspace*{-3ex}
\end{table}

In addition, 
our learned controller was evaluated against perturbations.
The robot was pushed at the base link frame for \unit[0.2]{s} from the front, the back and the side at different commanded velocities.
For each commanded velocity,
we started perturbing the robot with \unit[10]{N} pushes and increased this magnitude by \unit[10]{N} after ten pushes.
Table~\ref{table:perturbation_test} presents the maximum push the robot was able to reject successfully 10 times in a row (100\% Succ.), 
and the maximum push the robot was able to reject at least once (Max. push).

In order to evaluate the contribution of the velocity scheduler, 
we trained a policy without curriculum.
After the same number of epochs, 
the robot learned to stand but it was not able to walk in any direction. 
Additionally,
a controller was learned replacing the Beta policy by a Gaussian one.
With this controller, 
the robot was able to stand but it was not able to go more than four steps without falling due to constant saturation of the PD controllers.
This demonstrates that the introduction of Beta policies plays a key role on the use of energy and furthermore it avoids the incorporation of torque terms in the reward function and the corresponding weight assignment and torque measurements or estimates.

\begin{table}[]
	\centering
	\caption{Results of perturbation [N] test}
	\vspace*{-2ex}
	\label{table:perturbation_test}
	\begin{tabular}{cc|cccc}
		\toprule
		\textbf{Commanded}& \multirow{2}{*}{\textbf{Test}} & \textbf{Front} & \textbf{Back} & \multicolumn{2}{c}{\textbf{Lateral p.}}\\
		\textbf{velocity}& & \textbf{push} & \textbf{push} & right & left\\
		\midrule
		\multirow{2}{*}{In-place} & 100\% Succ. & 40 & 20 & \multicolumn{2}{c}{50} \\
		&Max. push & 60 & 30 & \multicolumn{2}{c} {90}\\
		\midrule
		\multirow{2}{*}{$v=[0.3, 0, 0]$} & 100\% Succ. & 30 & 20 & \multicolumn{2}{c}{50} \\
		&Max. push & 50 & 30 & \multicolumn{2}{c} {80}\\
		\midrule
		\multirow{2}{*}{$v=[0.6, 0, 0]$} & 100\% Succ. & 30 & 10 & \multicolumn{2}{c}{40} \\
		&Max. push & 50 & 20 & \multicolumn{2}{c}{70} \\
		\midrule
		\multirow{2}{*}{$v=[0, 0.25, 0]$} & 100\% Succ. & 40 & 20 & 40 & 40 \\
		&Max. push & 60 & 30 & 60 & 100\\
		\midrule
		\multirow{2}{*}{$v=[0, 0.5, 0]$} & 100\% Succ. & 40 & 20 & 20 & 50 \\
		&Max. push & 60 & 30 & 60 & 100 \\
		\bottomrule
	\end{tabular}
		\vspace*{-3ex}
\end{table}

\begin{figure}[b!]
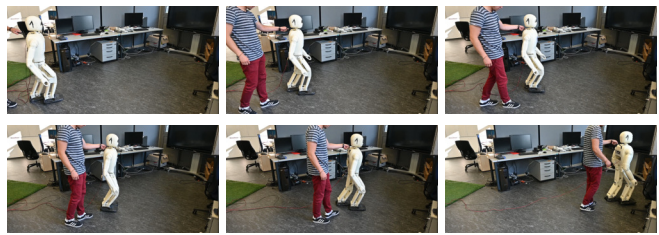

		\vspace*{-1ex}
	\centering
	\includegraphics[width=0.325\linewidth, trim={20pt 60pt 40pt 50pt}, clip]{imgs/transfer/forward/forward_2.png} \hfill
	\includegraphics[width=0.325\linewidth, trim={20pt 60pt 40pt 50pt}, clip]{imgs/transfer/forward/forward_6.png} \hfill
	\includegraphics[width=0.325\linewidth, trim={20pt 60pt 40pt 50pt}, clip]{imgs/transfer/forward/forward_7.png} \hfill
	\\ \vspace{1ex}
	\includegraphics[width=0.325\linewidth, trim={20pt 60pt 40pt 50pt}, clip]{imgs/transfer/forward/forward_8.png}
	\includegraphics[width=0.325\linewidth, trim={20pt 60pt 40pt 50pt}, clip]{imgs/transfer/forward/forward_9.png} \hfill
	\includegraphics[width=0.325\linewidth, trim={20pt 60pt 40pt 50pt}, clip]{imgs/transfer/forward/forward_12.png}
	\caption{
		Snapshots of forward walk performed by the real robot commanded by the learned locomotion controller.	}
	\label{fig:transfer_forward}
\end{figure}

\begin{figure}[]
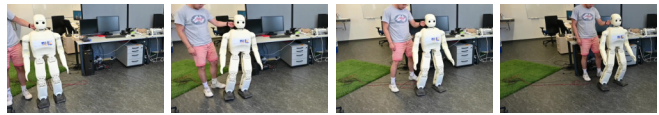

	\centering
	\vspace*{-1ex}
	\includegraphics[width=0.24\linewidth, trim={40pt 120pt 640pt 100pt}, clip]{imgs/transfer/left/image-001.jpeg} 
	\includegraphics[width=0.24\linewidth, trim={40pt 120pt 640pt 100pt}, clip]{imgs/transfer/left/image-018.jpeg} 
	\includegraphics[width=0.24\linewidth, trim={40pt 120pt 640pt 100pt}, clip]{imgs/transfer/left/image-032.jpeg}
	\includegraphics[width=0.24\linewidth, trim={40pt 120pt 640pt 100pt}, clip]{imgs/transfer/left/image-040.jpeg} 	
	\caption{
		Snapshots of lateral (left) walk performed by the real robot commanded by the learned locomotion controller.	}
	\label{fig:transfer_left}
	\vspace*{-2ex}
\end{figure}

Finally,
we transfer the learned gait to real hardware.
The NimbRo-OP2X robot is able to walk on the spot, forwards, backwards, laterally and diagonally.
Turning in place is also possible.
We observed dissimilarities between the simulated and the real gait which demands for more sophisticated sim-to-real transfer methods.
Snapshots of the robot walking forward and lateral are shown in Fig.~\ref{fig:transfer_forward} and Fig.~\ref{fig:transfer_left}. %
\section{Conclusion} 
\label{sec:conclusion}
We presented a novel approach to learn a single control policy capable of omnidirectional walking for humanoid robots using a realistic robot model.
We have demonstrated the capacity of the learned policy to walk in the sagittal and lateral directions and to turn around the vertical axis at different speeds.
Without altering the policy, 
our approach also produces motions in combined directions, i.e., the agent is able to walk diagonally and to turn during walking.
Achieving these locomotion behaviors was possible mainly due to:
the velocity scheduler,
the introduction of a core velocity;
the use of beta policies to bound the action space;
the incorporation into the state $s_t$ of the velocity of the base expressed in the relative and in the reference frame.
Our approach does not require reference motions to achieve anthropomorphic locomotion thanks to the introduction of a nominal pose.

In the future, we will explore different alternatives for motion regularizers to find one that is more flexible than motion capture data and generates more anthropomorphic motions compared to nominal poses.
In addition, 
we would like to extend this approach for 3D walking, 
such that the robot is able to walk on sloped terrains and to climb stairs. 

\balance
\printbibliography

\end{document}